\documentclass[a4paper]{article}

\usepackage{INTERSPEECH2016}
\usepackage{graphicx,mathtools,mathalfa,multirow,balance,url}
\usepackage{epsfig,epstopdf}
\usepackage{amssymb,amsmath,bm}
\usepackage{textcomp}
\usepackage{booktabs}

\ninept

\title{The IBM Speaker Recognition System: Recent Advances and Error Analysis}

\makeatletter
\def\name#1{\gdef\@name{#1\\}}
\makeatother \name{{\em Seyed Omid Sadjadi$^1$, Jason W. Pelecanos$^1$, Sriram Ganapathy$^{2}$}}

\address{$^1$IBM Research, Yorktown Heights, NY, USA \\
  $^2$Electrical Eng. Dept., Indian Institute of Science, Bangalore, India \\
  {\small \tt sadjadi@us.ibm.com}
}

\begin{document}

  \maketitle
  \begin{abstract}
  	We present the recent advances along with an error analysis of the IBM speaker recognition system for conversational speech. Some of the key advancements that contribute to our system include: a nearest-neighbor discriminant analysis (NDA) approach (as opposed to LDA) for intersession variability compensation in the i-vector space, the application of speaker and channel-adapted features derived from an automatic speech recognition (ASR) system for speaker recognition, and the use of a DNN acoustic model with a very large number of output units ($\sim\!\!\!10$k senones) to compute the frame-level soft alignments required in the i-vector estimation process. We evaluate these techniques on the NIST 2010 SRE extended core conditions (C1--C9), as well as the \textit{10sec--10sec} condition. To our knowledge, results achieved by our system represent the best performances published to date on these conditions. For example, on the extended \textit{tel-tel} condition (C5) the system achieves an EER of 0.59\%. To garner further understanding of the remaining errors (on C5), we examine the recordings associated with the low scoring target trials, where various issues are identified for the problematic recordings/trials. Interestingly, it is observed that correcting the pathological recordings not only improves the scores for the target trials but also for the non-target trials.
  \end{abstract}

  \section{Introduction}

    In recent years, the research trend in the speaker recognition field has evolved from joint factor analysis (JFA) based methods, which attempt to model the speaker and channel subspaces separately \cite{Kenny2007}, towards the i-vector approach that models both speaker and channel variabilities in a single low-dimensional (e.g., a few hundred) space termed the total variability subspace \cite{Dehak2011}. State-of-the-art i-vector based speaker recognition systems employ universal background models (UBM), which are based on either unsupervised GMMs \cite{Reynolds2000} or supervised ASR acoustic models (e.g., GMM-HMM or DNN) \cite{Omar2010, Lei2014a, Snyder2015}, to generate frame-level soft alignments required in the i-vector estimation process. DNN Bottleneck and Tandem features have also been explored for speaker recognition \cite{Heck2000,Yaman2012} , and more recently successfully used in some state-of-the-art i-vector systems \cite{Richardson2015, Matejka2016}. The i-vectors are typically post-processed through a linear discriminant analysis (LDA) \cite{Fukunaga1990} stage to generate dimensionality reduced and channel-compensated features which can then be efficiently modeled and scored with various backends such as a probabilistic LDA (PLDA) \cite{Prince2007, Garcia2011}. 
    
    In this paper, we report on the latest advancements made in the IBM i-vector speaker recognition system \cite{Sadjadi2016} for conversational speech. Particularly, we describe the key components that contribute significantly to our system performance. These components include: 1) a nearest-neighbor based discriminant analysis (NDA) approach \cite{Sadjadi2014} for channel compensation in the i-vector space, which, unlike the commonly used Fisher LDA, is non-parametric and typically of full rank, 2) speaker- and channel-adapted features derived from feature-space maximum likelihood linear regression (fMLLR) transforms \cite{Digalakis1995, Gales1998}, which are used both to train/evaluate the DNN and to compute the sufficient Baum-Welch statistics for i-vector extraction, and 3) a DNN acoustic model with a large number of output units ($\sim~\!\!\!10$k~senones) to compute the soft alignments (i.e., the posteriors). To quantify the contribution of these components, we evaluate our system in the context of speaker verification experiments using speech material from the NIST 2010 speaker recognition evaluation (SRE) which includes 9 extended core tasks as well as a \textit{10sec--10sec} condition. Motivated by the relatively low speaker recognition error rates achieved by our system (e.g., 0.59\% EER on C5 in SRE 2010), we also conduct an error analysis of low scoring target trials to gain insights regarding the nature of the issues associated with the remaining system errors on C5.

\begin{figure*}[t]
	\centering
	\includegraphics[scale=.48, clip, trim=0mm 8mm 0mm 0mm] {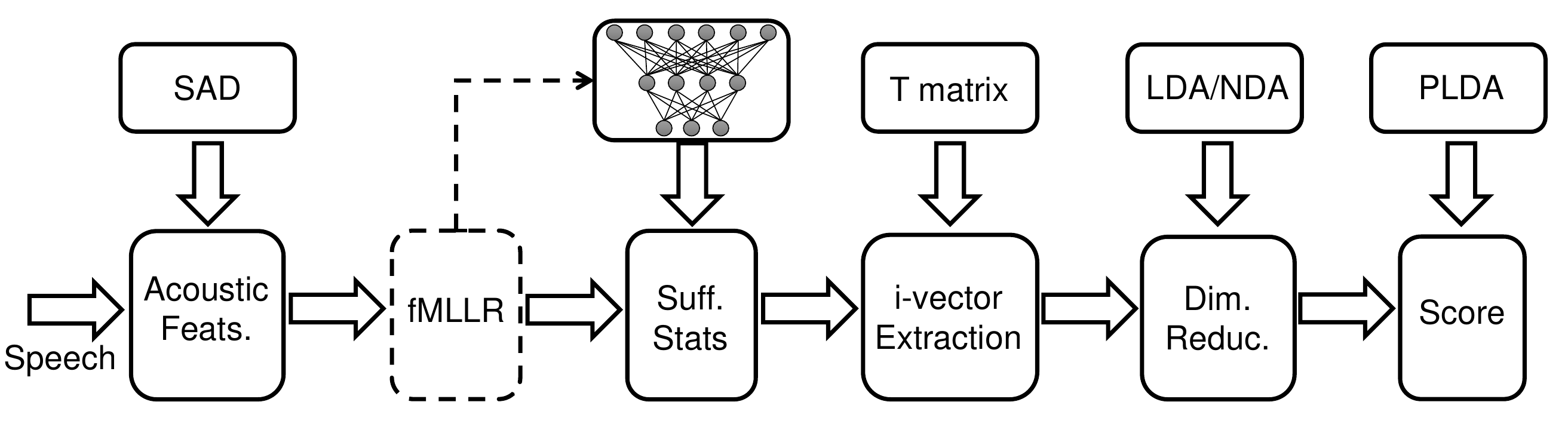}
	\vspace{-3mm}
	\caption{\it Block diagram of the IBM speaker recognition system with fMLLR speaker- and channel-adapted features, DNN posterior based i-vectors, and NDA dimensionality reduction.}
	\label{fig:blk}
	\vspace{-4mm}
\end{figure*}

\section{System Overview}

In the following subsections, we briefly describe the major components of our speaker recognition system. A schematic block diagram of the system is depicted in Fig.~\ref{fig:blk}.

\subsection{DNN i-vector extraction}
\vspace{-1mm}
The i-vector representation is based on the total variability modeling concept which assumes that speaker- and channel-dependent variabilities reside in the same low-dimensional subspace \cite{Dehak2011}. In order to learn the bases for the total variability subspace, one needs to first compute the Baum-Welch statistics which are defined as,\vspace{-1mm}
\begin{eqnarray}
\label{eqn:zeroth}
N_g(s) &=& \sum_{t}\gamma_{tg}(s),\\
\label{eqn:first}
\mathbf{F}_g(s) &=& \sum_{t}\gamma_{tg}(s)\, \mathbf{O}_t(s),
\vspace{-5mm}
\end{eqnarray}
where $N_g(s)$ and $\mathbf{F}_g(s)$ denote the zeroth- and first-order statistics for speech session $s$, respectively, with $\gamma_{tg}(s)$ being the posterior probability of the mixture component $g$ given the observation vector $\mathbf{O}_t(s)$ at time frame $t$. 

The observation vector $\mathbf{O}_t(s)$ can be either the conventional raw acoustic features such as MFCCs or their speaker- and channel-adapted versions which are computed through per recording fMLLR transforms \cite{Gales1998, Digalakis1995} typically obtained with a GMM-HMM system. Note from Fig.~\ref{fig:blk} that the same fMLLR transformed features can be used to train/evaluate the DNN as well as compute the Baum-Welch sufficient statistics for i-vector extraction.

Traditionally, the frame-level soft alignments, $\gamma_{tg}(s)$, in (\ref{eqn:zeroth}) and (\ref{eqn:first}) are computed with a GMM acoustic model trained in an unsupervised fashion (i.e., with no phonetic labels). However, in \cite{Omar2010}, a supervised GMM-HMM acoustic model (derived from a speech recognition system) was utilized to estimate the GMM-UBM hyperparameters for speaker recognition, assuming that class-conditional distributions for the various phonetic classes are Gaussian. More recently, inspired by the success of DNN acoustic models in the automatic speech recognition (ASR) field, \cite{Lei2014a} proposed the use of DNN senone (context-dependent triphones) posteriors for computing the soft alignments, $\gamma_{tg}(s)$, which resulted in remarkable reductions in speaker recognition error rates. Motivated by these results, in this paper, we explore DNN i-vectors extracted with a very large number of senones for speaker recognition, and compare their effectiveness against GMM i-vectors on this task.

\subsection{Nearest-neighbor discriminant analysis (NDA)}
\label{sec:nda}
\vspace{-1mm}
As noted previously, state-of-the-art speaker recognition systems employ LDA for intersession variability compensation in the i-vector space. However, there are some limitations associated with the parametric LDA where the underlying distribution of classes is assumed to be Gaussian and unimodal. Nevertheless, it is well known in the speaker recognition community that the actual distribution of i-vectors may not necessarily be Gaussian \cite{Kenny2010}, particularly in the presence of noise and channel distortions \cite{Sadjadi2014, Sadjadi2015}. In addition, for the NIST SRE scenarios, speech recordings come from various sources (sometimes out-of-domain), therefore unimodality of the distributions cannot be guaranteed.

In order to alleviate some of the limitations identified for LDA, a nonparametric nearest-neighbor based discriminant analysis technique was proposed in \cite{Fukunaga1983}, and recently evaluated for both speaker and language recognition tasks on high-frequency (HF) radio channel degraded data \cite{Sadjadi2014, Sadjadi2015} where it compared favorably to LDA. In NDA, the expected values that represent the global information about each class are replaced with local sample averages computed based on the $k$-NN of individual samples. More specifically, in the NDA approach, the between-class scatter matrix is defined as,\vspace{-2mm}
\begin{equation} \label{eqn:betweenk}
\tilde{\mathbf{S}}_b = \sum_{i=1}^{C} \sum_{\substack{j=1 \\ j\neq i}}^{C} \sum_{l=1}^{N_i} w^{ij}_l {\left(\mathbf{x}^i_l-\mathcal{M}^{ij}_l\right)\left(\mathbf{x}^i_l-\mathcal{M}^{ij}_l\right)^T},
\vspace{-3mm}
\end{equation}     
where $\mathbf{x}^i_l$ denotes the $l^\textrm{th}$ sample from class $i$, and $\mathcal{M}^{ij}_l$ is the local mean of $k$-NN samples for $\mathbf{x}^i_l$ from class $j$. Here, $C$ and $N_i$ denote the number of classes and the number of samples in class $i$, respectively. The weighting function $w^{ij}_l$ is introduced in (\ref{eqn:betweenk}) to deemphasize the local gradients that are large in magnitude to mitigate their influence on the scatter matrix. The weight parameter, $w^{ij}_l$, is larger for samples near the classification boundary, while it drops off to approximately $0$ for samples that are far from the boundary. In this study, the within-class scatter matrix, $\mathbf{S}_w$, is computed similarly as in LDA. The NDA transform is then formed by calculating the eigenvectors of $\mathbf{S}_w^{-1}\tilde{\mathbf{S}}_b$.

Three important observations can be made from a careful examination of the nonparametric between-class scatter matrix in (\ref{eqn:betweenk}). First, the mean vector, $\mathcal{M}^{ij}_l$, is calculated locally (as opposed to globally in LDA), which can result in more robust transforms, particularly for scenarios where unimodality of the class conditional distributions cannot be guaranteed. Second, because all the samples are taken into account for the calculation of the nonparametric between-class scatter matrix (as opposed to only the class centroids in LDA), $\tilde{\mathbf{S}}_b$, and hence the NDA projection, is generally of full rank. Finally, compared to LDA, NDA is more effective in preserving the complex structure (i.e., local structure) within and across different classes because LDA only uses global gradients obtained with the centroids of the classes to measure the between-class scatter. On the other hand, NDA uses local gradients that are emphasized along the boundary through the weighting function, $w^{ij}_l$.

 \section{Experiments}
 \label{sec:expts}
 
 This section provides a description of our experimental setup including speech data, the ASR system configuration, and the speaker recognition system configuration. 
  	
 	\subsection{Data}
 	\vspace{-1mm}
 	We conduct the core of our speaker recognition experiments using conversational telephone and microphone (phone call and interview) speech material extracted from datasets released through the LDC for the NIST 2004-2010 SREs \cite{Cieri2007, Brandschain2010}, as well as Switchboard Cellular (SWBCELL) Parts I and II and Switchboard2 (SWB2) Phases II and III corpora. These datasets contain speech spoken in U.S. English (the non English portion was filtered out) from a large number of male and female speakers with multiple sessions per speaker. The NIST SRE 2010 data is held out for evaluations, while the remaining data are used to train the system hyper-parameters (i.e., the i-vector extractor, LDA/NDA, and PLDA). In our experiments, we consider all 9 extended core tasks (C1--C9) along with the \textit{10sec--10sec} condition (C10sec) in the NIST SRE 2010 that involve telephone and microphone trials from both male and female speakers (consult \cite{SRE2010} for a more detailed description of the tasks).  
 	
 	\subsection{DNN system configuration}
 	\vspace{-1mm}
	 A DNN model, with 7 fully connected hidden layers with 2048 units per layer except for the bottleneck layer that has 512 units, is discriminatively trained using the standard error back-propagation and cross-entropy objective function to estimate posterior probabilities of 10,000 senones (HMM triphone states). The training is accomplished using the IBM Attila toolkit \cite{Soltau2010} on 600 hours of conversational telephone speech (CTS) data from the Fisher corpus \cite{Cieri2004} with a 9-frame context of 40-dimensional speaker-adapted feature vectors obtained using per recording fMLLR transforms \cite{Digalakis1995, Gales1998}. The fMLLR transforms are generated for each recording with decoding alignments obtained from a GMM-HMM acoustic model (see \cite{Ganapathy2015, Saon2015} for more details). 
 	
 	\subsection{Speaker recognition system configuration}
 	\vspace{-1mm}
 	For speech parameterization (other than the fMLLR based features), we extract 13-dimensional MFCCs (including $c_0$) from 25~ms frames every 10~ms using a 24-channel mel filterbank. The first and second temporal cepstral derivatives are also computed over a 5-frame window and appended to the static features to capture the dynamic pattern of speech over time. This results in 39-dimensional feature vectors. For non-speech frame dropping, we employ an unsupervised speech activity detector (SAD) based on voicing features \cite{Sadjadi2013}. After dropping the non-speech frames, short-time cepstral mean subtraction (CMS) is applied to suppress the short term linear channel effects.
 	
 	In this paper, a 500-dimensional total variability subspace is learned and used to extract i-vectors from the recordings.  To learn the i-vector extractor, out of a total of 60,178 recordings available from 1884 male and 2601 female speakers, we select 48,325 recordings from NIST SRE 2004-2008, SWBCELL, and SWB2 corpora. The zeroth and first order Baum-Welch statistics are computed for each recording using soft alignments obtained from either a gender-independent 2048-component GMM-UBM with diagonal covariance matrices, or the DNN acoustic model with 2k, 4k, and 10k senones. The GMM-UBM is trained using 21,207 recordings selected from the NIST SRE 2004-2006, SWBCELL, and SWB2 corpora.
 	
 	After extracting 500-dimensional i-vectors, we either use LDA or NDA for inter-session variability compensation by reducing the dimensionality to 250. In order to train the NDA, we employ a one-versus-rest strategy to compute the inter-speaker scatter matrix in (\ref{eqn:betweenk}). This provides flexibility on the number of nearest neighbors used for computing the local means. A cosine similarity metric (as opposed to Euclidean) is used to find the $k$-nearest neighbors for each sample. The dimensionality reduced i-vectors are then centered, whitened, and unit-length normalized. For scoring, a Gaussian PLDA model with a full covariance residual noise term \cite{Prince2007, Garcia2011} is learned using the i-vectors extracted from all 60,178 speech segments (1884 male and 2601 female speakers) as noted previously. The Eigenvoice subspace in the PLDA model is assumed full-rank.

 \section{Results and Discussion}
 \label{sec:result}
 
	\begin{table}[t]
		\vspace{-2mm}
		\renewcommand{\tabcolsep}{1.5 mm} 
		\renewcommand{\arraystretch}{1.2} 
		\caption{\label{tab:tab2} {\it Performance comparison of IBM speaker recognition systems with various configurations on C5, with 10k senones.}}
		\vspace{0mm}
		\centerline{
			\begin{tabular}{|l||c|c|c|}
				\hline
				System & EER [\%] & minDCF08 & minDCF10 \\
				\hline  \hline
				GMM-MFCC-LDA & 2.11  & 0.113 & 0.440 \\
				GMM-MFCC-NDA & 1.49 & 0.071 & 0.280 \\ \hline
				DNN-MFCC-LDA & 0.91 & 0.045 & 0.172 \\
				DNN-MFCC-NDA & 0.68 & 0.034 & 0.140 \\ \hline
				DNN-fMLLR-LDA & 0.75 & 0.032 & 0.125 \\
				DNN-fMLLR-NDA & \textbf{0.59} & \textbf{0.025} & \textbf{0.095} \\
				\hline
			\end{tabular}}
			\vspace{-2mm}
		\end{table}
		
		\begin{table}[b]
			\vspace{-7mm}
			\renewcommand{\tabcolsep}{1.2 mm} 
			\renewcommand{\arraystretch}{1.0} 
			\caption{\label{tab:tab3} {\it Comparison of LDA vs NDA for in-domain and out-of-domain training on C5, with fMLLR features and 10k senones.}}
			\vspace{0mm}
			\centerline{
				\begin{tabular}{|l|c|c|c|c|}
					\hline
					System & \text{Domain} & EER [\%] & minDCF08 & minDCF10 \\
					\hline  \hline
					& out & 1.80  & 0.085 & 0.256 \\
					DNN-LDA & in & 0.78 & 0.034 & 0.149 \\
					& in+out & 0.75 & 0.032 & 0.125 \\ \hline
					& out & 1.55 & 0.071 & 0.226 \\
					DNN-NDA & in & 0.75 & 0.033 & 0.119 \\
					& in+out & \textbf{0.59} & \textbf{0.025} & \textbf{0.095} \\
					\hline
				\end{tabular}}
				\vspace{-0mm}
			\end{table}		
 \subsection{Experimental results}
 \vspace{-2mm}
 In this section, we summarize the results obtained with the experimental setup presented in Section~\ref{sec:expts}. In the first experiment, we evaluated the effectiveness of NDA versus LDA for inter-session variability compensation in the i-vector space. The outcome of this experiment on C5 is presented in Table~\ref{tab:tab2}, in terms of the equal error rate (EER), minimum detection cost function with the NIST SRE 2008 \cite{SRE2008} and 2010 \cite{SRE2010} definitions (minDCF08 and minDCF10). It can be seen from the table that the systems with NDA consistently provide better speaker recognition performance across all three metrics. For the GMM based system, a relative improvement of ~30\% in EER is achieved with NDA over LDA, while for the DNN based systems with MFCCs and fMLLR features relative improvements of 25\% and 21\% are obtained, respectively. We speculate that this may be due to the nonparametric nature of the scatter matrices in NDA that makes no assumptions regarding the underlying class-conditional distributions (i.e., Gaussianity and unimodality). Another important observation that can be made from Table~\ref{tab:tab2} is that, irrespective of the dimensionality reduction algorithm used, the systems with fMLLR features outperform the MFCC based systems. This is possibly attributed to the ability of the fMLLR transforms in reducing the speaker and channel variabilities in the acoustic feature space. Finally, consistent with the results reported in recent studies \cite{Lei2014a, Snyder2015}, the DNN based systems outperform the GMM based systems by significant margin (e.g., resulting in a relative improvement of more than 54\% in terms of the EER with MFCCs and NDA). In \cite{Sadjadi2016}, we also investigated the impact of the senone set size (2k, 4k, and 10k) on speaker recognition performance, where we observed that the larger the number of senones, the better the performance. It is worth noting that increasing the number of components in the unsupervised GMM acoustic model (with diagonal covariance matrices) for speaker recognition did not seem to result in much performance gains in the recent studies \cite{Lei2014a, Snyder2015}. 
 
 \begin{table*}[t]
 	\renewcommand{\tabcolsep}{2.0 mm} 
 	\renewcommand{\arraystretch}{1.3} 
 	\caption{\it Performance comparison of IBM speaker recognition systems with various configurations on C1--C10sec (excluding C5).}
 	\eightpt
 	\vspace{-2mm}
 	\begin{center}
 		\begin{tabular}{ lccccccccc }
 			\hline
 			\multirow{2}{*}{System} & \multicolumn{9}{c}{EER [\%] (minDCF10)}\\ 
 			\cline{2-10}
 			{ } & {C1} & {C2} & {C3} & {C4} & {C6} & {C7} & {C8} & {C9} & {C10sec}\\
 			\hline
 			{GMM-MFCC-NDA} & 1.2 (0.22) & 1.8 (0.32) & 1.8 (0.30) & 1.1 (0.25) & 2.9 (0.55) & 3.3 (0.59) & 1.1 (0.25) & 0.7 (0.11) & 11.7 (0.99)\\
 			{DNN-MFCC-NDA} & \textbf{0.8 (0.11)} & \textbf{1.3} (\textbf{0.16}) & 0.9 (0.13) & \textbf{0.6} (0.11) & 1.5 (0.34) & \textbf{1.1} (0.32) & 0.6 (0.14) &  0.4 \textbf{(0.05)} & \textbf{8.3 (0.94)} \\
 			{DNN-fMLLR-NDA} & 0.9 (0.13) & \textbf{1.3} (0.17) & \textbf{0.9 (0.10)} & 0.7 (\textbf{0.08}) & \textbf{1.0 (0.24)} & 1.4 \textbf{(0.31)} & \textbf{0.4 (0.10)} & \textbf{0.3 (0.05)} & 12.3 (0.99) \\
 			\hline
 		\end{tabular}
 	\end{center}
 	\label{tab:tab4}
 	\vspace{-5mm}
 \end{table*}     	
 
 In the next set of experiments, we investigated the impact of in-domain and out-of-domain training for LDA versus NDA. This was accomplished by first splitting the training data into in- and out-of-domain parts, and then retraining the LDA and NDA models on these parts. Similar to the setup used in the domain adaptation challenge (e.g., see \cite{Richardson2015, Garcia2014}), the in-domain part contains CTS data from the \textit{Mixer} collection, i.e., SRE2004--2008 corpora, while the out-of-domain part contains data from the older SWB2 and SWBCELL corpora. Table~\ref{tab:tab3} shows the results from these experiments on C5 which are obtained with i-vectors computed using 10k DNN senones and fMLLR features. It is observed that while NDA outperforms LDA for out-of-domain training, it offers minimal improvement (at least in terms of EER) when only in-domain training data is used. However, with pooled in-domain and out-of-domain data (which is no longer unimodal), a significant improvement in performance is obtained with NDA over LDA (i.e., 21\% relative in terms of EER). This improvement may be attributed to the robustness of NDA to multimodal data, as discussed in Section~\ref{sec:nda}.

 For completeness, we also evaluated the performance of our speaker recognition system on extended microphone and telephone tasks, under normal and high/low vocal effort, (C1--C9) as well as the \textit{10sec--10sec} condition in the NIST SRE 2010. The results are provided in Table~\ref{tab:tab4} for both the GMM and DNN based systems. It is clear that the DNN based systems, with either MFCCs or fMLLR features, perform significantly better than the GMM based system. Additionally, the DNN based system trained with raw MFCCs tends to perform better than the fMLLR based system on interview microphone conditions (C1--C2). We speculate that this is because the fMLLR transforms, which are obtained using GMM-HMMs trained only on telephony data, are unable to effectively reduce the variability due to channel mismatch on microphone recordings. For other tasks, overall, MFCC and fMLLR based systems perform similarly except for the C10sec where, among other challenges, the VTL normalizations and the fMLLR transforms are also adversely impacted by the short duration of recordings (i.e., 10~s).     
 
\subsection{Error analysis}
\vspace{-1mm}
In consideration of the obtained results, for example 0.59\% EER on the extended condition 5, we attempt to gain an understanding of some of the underlying issues in the system. In particular, we would like to examine the characteristics of the recordings contributing to some of the lowest scoring target trials. 
Towards this, we analyzed 265 recordings relating to the 200 lowest scoring target trials and attempted to characterize them with various potentially problematic properties. It was found that, while many of the recordings are audibly acceptable, there are 76 with co-channel speech (either through cross-channel feedback or from background competing speakers), 51 with background noise, music, breath, sniffle, and fidgeting sounds, 33 with very sparse speech activity, 8 with cocktail party noise, and 2 exhibiting signal clipping effects.

Our objective is to improve the target trial scores relating to the problematic files. In particular, we focused our attention on three categories: (i) cocktail party effect, (ii) general noise, and (iii) co-channel speaker. These categories were chosen because they may benefit from manually changing the SAD information to more appropriately reflect the speech portions of the speaker of interest (i.e., the target speaker).

\begin{figure}[t]
	\centering
	\includegraphics[scale=0.495, clip, trim=0mm 0mm 0mm 0mm]{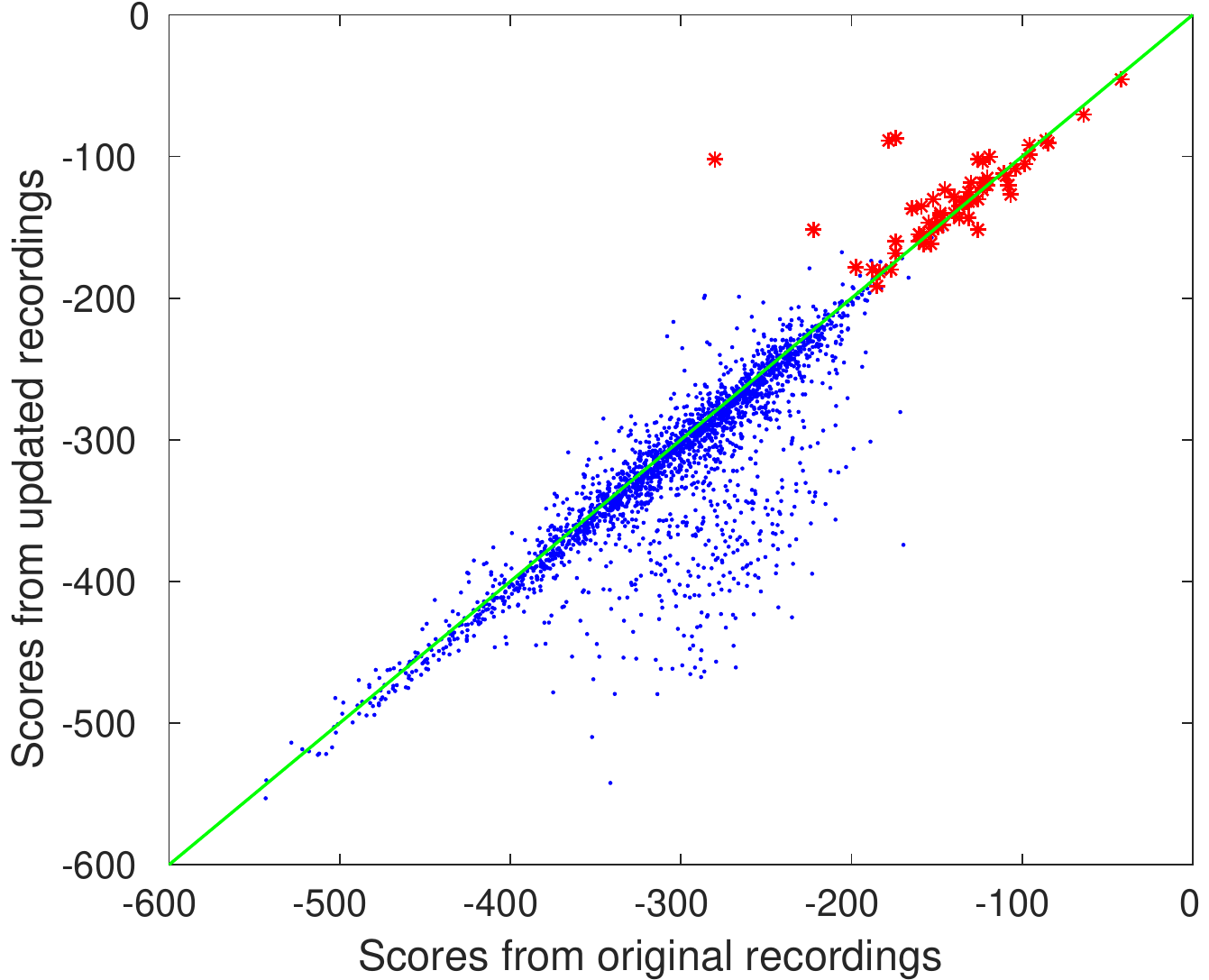}
	\vspace{-2mm}
	\caption{{\it Scatter plot of the scores for repaired trials versus the original trials. The asterisks ($*$) represent the target trials while the dots ($\cdot$) represent the non-target trials.}}
	\label{fig:scatterplot}
	\vspace{-2.5mm}
\end{figure}

Figure~\ref{fig:scatterplot} shows the result of the manual modification of the SAD labels by plotting the trial scores (both target and non-target) that related to a selection of 18 modified SAD recordings. The x-axis represents the scores from trials with the original recordings while the y-axis shows the scores for the modified recordings. An interesting point is that not only did some of the target scores significantly improve, but many of the non-target scores also decreased. One view is that the presence of a co-channel speaker has the effect of bringing the resulting i-vector closer to the general population of speaker i-vectors (on average). This raises an important observation that not only should a speech detection component be utilized but also a component that only focuses on the speaker of interest (or the homogeneity of the class). The large movement in both target and non-target trials was basically described by 8 of the 18 recordings. These consisted of intermittent noise conditions, cocktail party noise, and co-channel (cross-channel) speech where the interference was significant or comparable to the level of the target speech. Interestingly, by \textit{manually} modifying the SAD for 18 recordings the EER is reduced from 0.59\% to 0.56\%.

Future work will involve the automation of such manual processes, perhaps through the use of speaker diarization and/or audio enhancement (e.g. see \cite{Plchot2016}), as well as the expansion of the types of issues addressed in the system such as short-duration modeling~\cite{Kenny2013} and signal quantization issues.

\section{Conclusions}
In this paper, we presented the recent advancement made in our state-of-the-art i-vector speaker recognition system. We investigated the impact of several key components of the system on performance using extended core tasks in the NIST 2010 SRE that involved both microphone and telephone trials. Some important observations were as follows: 1) the NDA was found to be more robust than LDA to multimodality within data, hence more effective than LDA for inter-session variability compensation, 2) the fMLLR based features provided a better representation than raw MFCCs for matched data conditions (i.e., telephony trials), and 3) the DNN based UBM with a very large number of components (i.e., 10k senones) resulted in remarkable improvements in the performance of our system. Motivated by the relatively low EER achieved by our system (e.g., 0.59\% on C5), we also conducted an error analysis of low scoring target trials in C5 that revealed various issues (such as co-channel speech, background noise/music, and signal clipping effects) in the recordings associated with these problematic trials.  

  \newpage
  \eightpt
  \balance
  \bibliographystyle{IEEEtran}
  \bibliography{refs_is2016,IEEEabrv,IEEEfull}

\end{document}